\documentclass[11pt]{article}

\usepackage{geometry}
\usepackage[utf8]{inputenc} 
\usepackage[T1]{fontenc}    
\usepackage{hyperref}       
\usepackage{url}            
\usepackage{booktabs}       
\usepackage{amsfonts}       
\usepackage{nicefrac}       
\usepackage{microtype}      
\usepackage{xcolor}         
\usepackage{graphicx}
\usepackage{algorithmic}
\usepackage{algorithm2e}
\usepackage{amsmath}
\RestyleAlgo{ruled}
\SetKwComment{Comment}{/* }{ */}
\newtheorem{definition}{Definition}

\usepackage{multirow}
\newtheorem{Theorem}{Theorem}

\title{Density based Spatial Clustering of Lines via Probabilistic Generation of Neighbourhood}

\author{%
  Akanksha Das and Malay Bhattacharyya \\
  Indian Statistical Institute \\
  Kolkata -- 700108 \\
  India \\
  E-mail: akankshadas2020@gmail.com, malaybhattacharyya@isical.ac.in \\
}

\begin{document}

\maketitle

\begin{abstract}
Density based spatial clustering of points in $\mathbb{R}^n$ has a myriad of applications in a variety of industries. We generalise this problem to the density based clustering of lines in high-dimensional spaces, keeping in mind there exists no valid distance measure that follows the triangle inequality for lines. In this paper, we design a clustering algorithm that generates a customised neighbourhood for a line of a fixed volume (given as a parameter), based on an optional parameter as a continuous probability density function. This algorithm is not sensitive to the outliers and can effectively identify the noise in the data using a cardinality parameter. One of the pivotal applications of this algorithm is clustering data points in $\mathbb{R}^n$ with missing entries, while utilising the domain knowledge of the respective data. In particular, the proposed algorithm is able to cluster $n$-dimensional data points that contain at least $(n-1)$-dimensional information. We illustrate the neighbourhoods for the standard probability distributions with continuous probability density functions and demonstrate the effectiveness of our algorithm on various synthetic and real-world datasets (e.g., rail and road networks). The experimental results also highlight its application in clustering incomplete data.
\end{abstract}

\section{Introduction}
To understand the underlying structure of data, it is customary to explore groups within it [1]. The approach of identifying groups in data through unsupervised learning techniques is popularly known as clustering. Clustering of data points has received enormous attention in the last few decades. However recently, there is an emerging interest to generalize the clustering of data points to other shapes, particularly lines [2, 3]. The state-of-the-art clustering methods for points can not be directly extended to other shapes. The major challenges behind addressing this new class of problems are defining an appropriate distance measure and interpretation of clustering results.

There are many different approaches of clustering that perform well only for a specific type of datasets. For example, centroid-based clustering methods work well only for convex datasets [4]. The clustering of straight lines has become an appealing problem in recent years [2, 3, 5]. Unfortunately, this problem has been addressed with centroid-based approaches only. These methods suffer from data in which varying density of clusters is a concern. In density based clustering, groups are formed in a data space covering a continuous area with a high point density, divided from other groups by sparse regions. Density based spatial clustering of points in $\mathbb{R}^n$ has a myriad of applications in a variety of industries. In this paper, we propose a density based spatial clustering approach (termed as DeLi) of lines via probabilistic generation of neighbourhood aiming to cluster lines. The proposed approach is inspired from a density-based spatial clustering approach that effectively deals with noise (popularly known as DBSCAN) [6].

We address the problem of clustering of lines (or line segments) in a generic form. Unlike the conventional scenarios of clustering of lines, where every point on the same line get an equal preference, we make it flexible by introducing the concept of probability. We consider datasets consisting of both points and lines to demonstrate the effectiveness of our approach. Instead of considering a distance measure between a pair of lines, we adopt a novel concept of probabilistic neighbourhood generation to measure the proximity of those lines. In this way, our contributions in this area are indeed novel.

The rest of the paper is organized as follows. Section~2 discusses the major motivations behind the current work. Section~3 covers the state-of-the-art. Section~4 introduces the precursory notations and terminologies and formulates the problem. Section~5 presents the theoretical insights and details on the proposed approach of clustering of lines. Section~6 gives an overview of the datasets and presents the detailed experimental results. Finally, Section~6 concludes the paper by highlighting its advantages, disadvantages, and future prospects.

\section{Motivation}
There are several applications highlighted in the literature related to clustering of lines [2, 3]. However, in this paper, we bring the concept of density around a line envisioning its broader scope. This helps us to look at the problem of clustering in a generic way to foresee additional applications. Grouping of incomplete data is the major motivation behind the current work [7]. Let there be a set of points in $\mathbb{R}^n$ that we have to cluster, even though some of them might possess incomplete entries (i.e., values are not available for all the $n$-given dimensions). Mathematically speaking, say a data item $D \in \mathbb{R}^n$ have missing values on $k$ of its entries such that $k << n$. To address this version of clustering problem [8], we may consider the set of possible values of $D$ as a $(n-k)$ dimensional object in $\mathbb{R}^n$. In this study, we particularly deal with the scenarios where $k = 1$, i.e., we visualise such incomplete information as lines or line segments, and cluster points and line segments together under the same approach.

\section{Related Work}
The very first attempt to formulate clustering of lines was made by Peretz in the form of finding k-median for lines back in 2011 [5]. Given a set of lines $L$, he aimed to find $k$ points (facilities) such that the sum of distance from each line in $L$ to its closest point is minimized. Peretz proposed both exact and approximate algorithms for the said problem. It was also shown that $k$-median for lines problem can be solved efficiently for a constant $k$, however is NP-complete if $k$ is non-constant. Unfortunately, this approach works only well for convex datasets and for the cases where density is not varying. Moreover, it cannot identify the outliers in the data. Later Moram et al. have addressed the problem of clustering of lines with another centroid-based approach, precisely $k$-means in 2019 [3]. This also suffers from similar kind of limitations. We propose a density-based approach to overcome these limitations. Moreover, in recent times, grouping of a set of points with varying amount of reliability has been addressed by probabilistic clustering approaches for various applications [9, 10]. In this paper, we adopt a probabilistic approach in this paper to set up the neighbourhood of a dense cluster.

\section{Proposed Problem}
We first introduce the notations that will be used throughout the paper followed by the statement of the problem that we address.

\subsection{Basic Notations}
Let us define $\mathcal{U}$ as the set of all lines or line segments in $\mathbb{R}^n$ that we want to cluster. A line $l \in \mathbb{R}^n$ is denoted by the coordinates of any two distinct points on the line. On the other hand, a line segment is denoted by the coordinates of its end points, i.e., $l \in \mathbb{R}^n$ is represented by $(x, y)$, where $x, y \in \mathbb{R}^n$. Let us assume that a point $P$ on $l$ is denoted as $P \in l$. We consider $\ell^2$-norm as the distance measure between any pair of points, i.e., $\mathcal{D} (x, y) = ||x - y||_2$, where $x, y \in \mathbb{R}^n$. Let $d_{P, l}$ denote the shortest distance from $P$ to $l$, i.e., $d_{P, l} = \underset{Q \in l}{inf}\mathcal{D}(Q, P)$. Let us also define $P_l = Q \in l$ such that $\mathcal{D}(Q, P) = d_{P,l}$. 

\begin{Theorem}$P_l$ exists and is unique with respect to the line or line segment $l$.
\end{Theorem}
\textbf{proof :} We consider the two cases based on whether $l=(x,y)$ is a line or line segment.
$\square$
\begin{description}
\item[Case 1:] If $l$ is a line, $d_{P,l}$ is the perpendicular distance from $P$ to $l$ hence, $P_l$ is the foot of the perpendicular from $P$ to $l$. Proof is immediate.\\
\item[Case 2:] If $l$ is a line segment, and foot of the perpendicular(say $F$) lies on $l$ then same as in Case 1. If not, then both $x$ and $y$ lie on the same side of $F$ and $x,y$ and $ F$ are co linear. Wlog, let $x$ lie closer to $F$ than $y$ then $d_{P,l}=\mathcal{D}(x,P)$ and $P_l=x$. This is because $\mathbf{D}(P,z)$ is a strictly increasing function as $z$ varies from $x$ to $y$.\\
\end{description}
 
 Let $V(A)$ denote the volume of region $A \subseteq \mathbb{R}^n$ if it exists. The other notations are standard unless specified otherwise.

\subsection{Problem Formulation}
Let us denote the continuous probability density function as $f_l : L \rightarrow \mathbb{R}^+ \cup \{0\}$, where $L = \{P : P \in l\}$ is taken as a parameter towards formulating the problem in a generic way. Let $S_l$ denote the closure of $\{P \in l : f_l(P) > 0\}$. Based on this, we now define the neighbourhood of a line as follows.

\begin{definition}[$f$-neighbourhood of a line]
The $f$-neighbourhood of $l$ is defined as  $N_{f, l} = \{P\in \mathbb{R}^n : d_{P,l} < f(P_l) \}$, where $f: L \rightarrow \mathbb{R}^+ \cup \{0\}$ is any continuous function and $L = \{P : P \in l\}$.
\end{definition}

We now introduce a scaling factor for the normalization of the $f$-neighbourhood. This scaling factor can either be learned from the data (precisely from the volume parameter) or taken as a free parameter (in the absence of volume parameter). For the former case, the scaling factor is defined as follows.

\begin{Theorem}
Volume of $f$-neighbourhood of a line/line segment $l$  always exists where $f: L \rightarrow \mathbb{R}^+ \cup \{0\}$ is any continuous probability density function and $L = \{P : P \in l\}$.
\end{Theorem}

\textbf{proof:} Now $f$ is continuous, bounded and integrable over $L$ and $f$-neighbourhood of $l$ is the region bounded by rotating $f$ $360^o$ about $l$, and $l$. Hence, $V(N_{f,l})$ exists.  
$\square$

\begin{definition}[Scaling factor of $f$-neighbourhood of a line]
The scaling factor of $f$-neighbourhood of $l$ is defined as $\alpha_l = \frac{V}{V(N_{f_l,l})}$, where $V$ is the volume parameter.
\end{definition}

Using this scaling factor, we finally define the neighbourhood of a line as follows.

\begin{definition}[Neighbourhood of a line]
The \textit{neighbourhood} of $l$ is defined as $N_l = N_{(\alpha_l * f_l), l }$, i.e., the $(\alpha_l * f_l)$-neighbourhood of a line.
\end{definition}

Now, we define the neighbourhood relation $\mathcal{R}$ over $\mathcal{U}$, to determine whether $l_1$ is in the neighbourhood of $l_2$. Neighbourhood relation $\mathcal{R}$ is defined using neighbourhood of a line or using a suitable distance metric in the absence of $f_l$ parameter.

\begin{definition}[Neighbourhood Relation] Neighbourhood relation is defined as
\begin{equation*}
  l_1\mathcal{R}l_2 = \begin{cases}
    \exists P \in S_{l_2}$ such that $P \in N_{l_1}, & \text{if $f_l$ is available as a parameter}.\\
    \underset{p\in l_1,q\in l_2}{inf} \mathcal{D}(p,q)$ < $\alpha_{l_1}, & \text{otherwise}.
  \end{cases}
\end{equation*}
\end{definition}

Note that $\mathcal{R}$ is reflexive but not symmetric and transitive. We define the $l$-cardinality to verify whether a sufficient number of lines (a lower bound) is included within a cluster. This is defined as follows.

\begin{definition}[$l$-cardinality]
We define $l$-cardinality as the cardinality of the set $\{l_1: l_1\in \mathcal{U}$ and $l\mathcal{R}l_1\}$
\end{definition}

Given all the above precursory notations and definitions, we formulate the target problem as follows.

\textbf{Problem Statement:} \texttt{
Given $\mathcal{U}$ (a set of lines / line segments in $\mathbb{R}^n$), and a set of parameters as listed below:
\begin{itemize}
    \item Compulsory parameter: $c$ (cardinality),
    \item Substitute Parameters : $V$ (volume) or $\alpha_l$ (scaling factor), and
    \item Optional parameter : $f_l$,
\end{itemize}
we need to find out density based clusters of the lines / line segments in $\mathcal{U}$.
}

\section{Proposed Method}
We first put forward some theoretical aspects of the function $f_l$, to elaborate on the computation of the neighbourhood relation $\mathcal{R}$ and then present the algorithm for density based spatial clustering of lines / line segments.

\subsection{Theoretical Insights}
Clustering a set of lines by the existing approaches assumes all the points on a particular line / line segment to have equal importance [9]. However, in some real-world applications, it may be useful to assume that they have varying amounts of reliability. Hence, we assign a probability density to each of the points on a line / line segment and generate a neighbourhood accordingly. For example, if $p_1$ and $p_2$ are points lying on $l$ such that $f_l(p_1) >> f_l(p_2)$, then the span of the neighbourhood of $l$ at $p_1$ will be much larger than the span of the neighbourhood of $l$ at $p_2$. Assigning a probability density, gives us the autonomy to quantify the importance or reliability of the points on the line. This plays a vital role in incorporating domain knowledge of the variable with missing values during the clustering of incomplete data. 

The neighbourhoods for the standard probability distributions are shown in Table~\ref{Table:Neighbourhood}. As can be seen from Table~\ref{Table:Neighbourhood}, the neighbourhoods in $\mathbb{R}^3$ are simple extensions of what we have as neighbourhoods in $\mathbb{R}^2$. To illustrate this in depth, it is beneficial to transform the domain of $f_l$ from $\{P: P\in l\}$ to a subset of $\mathbb{R}$. This makes it easier to define $f_l$, thereby making the implementation simpler too. For a line segment $l = (x, y)$, where $x, y \in \mathbb{R}^n$, the set of points belonging to $l = (x, y)$  can be written as $\{x + (y - x)t : t \in [0, 1]\}$. Similarly, the set of points belonging to a line can be denoted as $\{x + (y - x) t : t \in \mathbb{R}\}$. As there is a bijection, more precisely a linear function, say $g_l$ \label{g_l} such that $g_l(0) = x, g_l(1) = y$  between the value $t$ and the points on the line or line segment, $f_l$ may be re-written as a function of $t \in [0, 1]$ (for line segments) or $\mathbb{R}$ (for lines) respectively. In Table~\ref{Table:Neighbourhood}, we define $f_l$ as a function of $t$ for the standard probability distributions, and show the region $N_l$ in $\mathbb{R}^2$ and $\mathbb{R}^3$. For each of the cases in Table~\ref{Table:Neighbourhood}, $f_l$ is defined over $\mathbb{R}$.

\begin{table}[htp]
  \caption{Generation of neighbourhoods in $\mathbb{R}^2$ and $\mathbb{R}^3$ using different standard probability distributions around a line. The scaling factor is considered to be 1.}
  \label{Table:Neighbourhood}
  \centering
  \begin{tabular}{lccc}
    \toprule
    \cmidrule(l){1-3}
    \textbf{Distribution} & \textbf{Parameters} & \textbf{2-D Neighbourhood} & \textbf{3-D Neighbourhood} \\
    \midrule
    Uniform & $(a,b)$ & \includegraphics[height=0.3in]{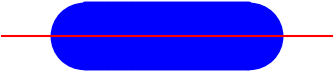} & \includegraphics[height=0.5in]{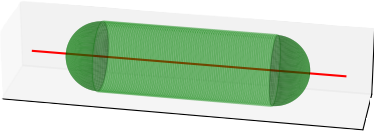} \\\hline
    Normal & $(\mu,\sigma^2)$ & \includegraphics[height=0.6in]{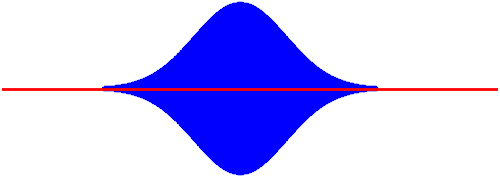} & \includegraphics[height=0.7in,width=1.2in]{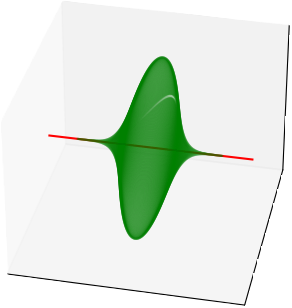} \\\hline
    Ellipsoidal & $(a,b)$ & \includegraphics[height=0.4in]{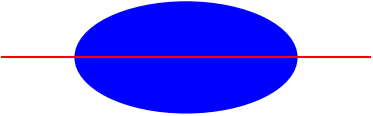} & \includegraphics[height=0.5in]{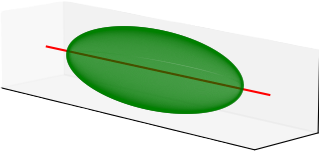} \\\hline
    Gamma & $(\alpha,\lambda)$ & \includegraphics[height=0.4in]{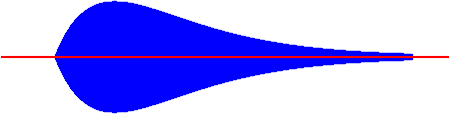} & \includegraphics[height=0.7in,width=1.2in]{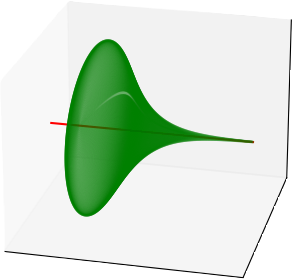} \\\hline
    Beta & $(\alpha_1,\alpha_2)$ & \includegraphics[height=0.6in]{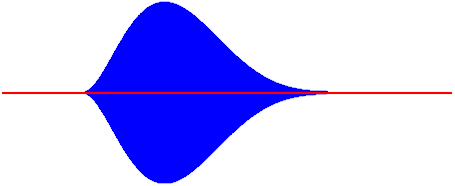} & \includegraphics[height=0.7in,width=1.2in]{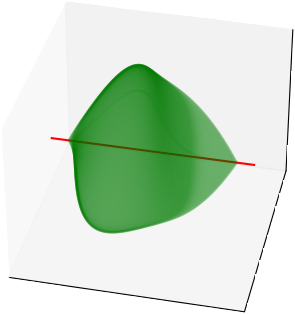} \\\hline
    Exponential & $(\lambda)$ & \includegraphics[height=0.5in]{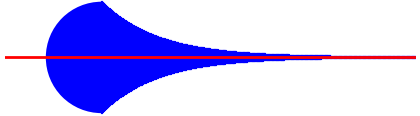} & \includegraphics[height=0.6in]{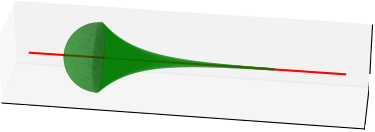} \\
    \bottomrule
  \end{tabular}
\end{table}

\subsection{Algorithm}
The proposed density based clustering algorithm of lines (termed as DeLi) has three versions differing only in the definition of the neighbourhood relation $\mathcal{R}$. Hence, the definitions of  $\mathcal{R}$ along with their associated parameters is concisely presented in Table~\ref{Table:Versions}.

\begin{table}[htp]
  \caption{Different versions of the algorithm for the computation of $l_1\mathcal{R}l_2$ and the respective output based on the input parameters.}
  \label{Table:Versions}
  \centering
  \begin{tabular}{ccccc}
    \toprule
    \cmidrule(l){1-3}
    \textbf{Version} & \multicolumn{3}{c}{\textbf{Parameters}} & \textbf{Computation of $l_1\mathcal{R}l_2$}
    \\\cline{2-4}
     & \textbf{Mandatory} & \textbf{Substitute} & \textbf{Optional} & (see Def.~4) \\
    \midrule
    1 & $c$ & $\alpha_l$ & - & $\underset{p\in l_1,q\in l_2}{inf} \mathcal{D}(p,q)$ < $\alpha_{l_1}$ \\
    2 & $c$ & $V$ & $f_l$ & $\exists P \in S_{l_2}$ such that $P \in N_{l_1}$ \\
    3 & $c$ & $\alpha_l$ & $f_l$ & $\exists P \in S_{l_2}$ such that $P \in N_{l_1}$ \\
    \bottomrule
  \end{tabular}
\end{table}

\begin{algorithm}[htp]
\caption{DeLi: Density based spatial clustering of lines via probabilistic generation of neighbourhood}
\label{Algorithm:1}
\KwIn{A set of lines or line segments $\mathcal{U} = \{l_1, l_2, \ldots, l_n\}$, a cardinality parameter $c$.\\}
\KwOut{The clusters $C_1, C_2, \ldots, C_k$.}
\textbf{Steps of the algorithm:}\\
\For{$l_i \in \mathcal{U}$}{
     Label $l_i$ as \texttt{UNVISITED} 
}
$j \leftarrow 1$ \\
\While{At least one $l_i \in \mathcal{U}$ is \texttt{UNVISITED}}{
$l_U \gets$ A randomly chosen $l_i$ that is labeled as \texttt{UNVISITED} \\
$N_U \gets \{l_i \in \mathcal{U}: l_U\mathcal{R} l_i\}$ \Comment{Computation of $l_U\mathcal{R} l_i$ is elaborated in Table~\ref{Table:Versions}}
\eIf{$|N_U| \geq c$}{
     Label $l_U$ and the lines or line segments in $N_U$ as \texttt{VISITED} \\
     Include $l_U$ and the lines or line segments in $N_U$ in a new cluster $C_j$ \\
     $j \leftarrow j + 1$ \\
}{
     Label $l_U$ as \texttt{NOISE}
}
}
\Return{$C_1, C_2, \ldots, C_k$}
\end{algorithm}

We formally present the approach in Algorithm~\ref{Algorithm:1}. The input to the algorithm is a set of lines / line segments in $\mathcal{U}$ and a set of appropriate parameters in accordance with the desired version (refer to Table~\ref{Table:Versions}). The algorithm closely mirrors the DBSCAN algorithm [6] for points with the neighbourhood being defined with the help of a probability density function $f_l$ instead of a metric. All lines / line segments are declared unvisited. If at least one line / line segment is unvisited, choose any one of them, say $l_U$ and run a loop over all lines / line segments to check their inclusions in the neighbourhood. If $l_U\mathcal{R}l_i$ holds (refer to Table~\ref{Table:Versions}), then add $l_i$ to the list $N_U$. To verify whether $l_U\mathcal{R}l_i$ holds, we choose one of the following versions based on the input parameters.

    \begin{description}
        \item[Version 1:] The minimum distance between $l_U = (x_U,y_U)$ and $l_i=(x_i,y_i)$ is strictly less then than $\alpha_U$ i.e  $\underset{x \in l_i, y \in l_U}{min}||x-y||_2^2 < \alpha_U^2$ that is equivalent to  $\underset{t_1\in D_1, t_2\in D_2}{min}||g_{l_i}(t_1)-g_{l_U}(t_2)||_2^2 < \alpha_U^2$, where $D_1, D_2 = \mathbb{R}$ or $[0,1]$. \ref{g_l}. Now $g_{l_i},g_{l_U}$ are linear functions hence the problem reduces to finding out the minimum value of a quadratic equation in two variables.
        
        \item[Version 2:] Calculate the $V(N_{l_U})$ and subsequently $\alpha_{l_U}=\frac{V}{V(N_{l_U})}$. Now, if any point in $l_i$ lies in the $(\alpha_{l_U}*f_{l_U})$-neighbourhood of $l_U$, then $l_U\mathcal{R}l_i$.

        \item[Version 3:] If any point in $l_i$ lies in the $(\alpha_{l_U}*f_{l_U})$-neighbourhood of $l_U$, then $l_U\mathcal{R}l_i$.
    \end{description}

If cardinality of $N_U$ is at least $c$, the $l_U$ and the elements of  $N_U$ are labeled as visited and are included in a new cluster $C_j$ where $j$ is auto-incremented. Otherwise, $l_U$ is labeled as noise i.e an outlier. Repeat until no line/line segment is left unvisited.
Finally the clusters $C_1, C_2, \ldots C_k$ are returned as the result.

\subsection{Complexity Analysis}
The time and space requirements of the proposed algorithm are discussed below.

\textbf{Time complexity:} The initial labelling in the proposed algorithm requires $O(n)$ time, for $n$ lines / line segments. Thereafter, the lines / line segments are processed one at a time and for each of them relation to the others are computed, incurring $O(n^2)$ time. Hence, the total worst case time complexity becomes $O(n) + O(n^2) \equiv O(n^2)$.

\textbf{Space complexity:} The proposed algorithm computes $\mathcal{R}$, which is not symmetric. Hence, we need not store a matrix representation of computations of neighbourhood within an auxiliary space to avoid re-computations during the runtime. So, for the $n$ lines / line segments (each represented as a doublet) to be clustered, the worst case space complexity is $O(n)$.

\section{Results}
To test the effectiveness of the proposed approach, we have applied it on various types of datasets. We first detail on the datasets used and then discuss the experimental results obtained.

\subsection{Dataset Details}
The datasets considered for experimental analysis were of two types -- Line Datasets and Point Datasets with Missing Entries. The Line Datasets are further categorised as Synthetic Datasets and Real-world Datasets. The details about the datasets are provided below (see Table~\ref{Table:Datasets} for more details).

\begin{itemize}
    \item \textbf{Line Datasets:} These datasets include line segments and are of two types. No domain knowledge is used for this kind of datasets.
    \begin{itemize}
        \item \textbf{Synthetic:} These are prepared by writing programs with carefully controlled parameters. There are two such datasets, namely Convex and Doughnut.
        \item \textbf{Real-world:} These are prepared by selecting a portion of real-world datasets to create clusters recognisable through human perception to test the efficacy of our algorithm. There are six such datasets namely Sparse Tripod, Dense Tripod, Broken Beads 1, Broken Beads 2, Spectral Band 1, and Spectral Band 2, which include Amtrak routes and subway lines in USA [11].
    \end{itemize}
    \item \textbf{Point Dataset with Missing Entries:} This dataset includes points with missing entries and are of one type. We consider domain knowledge for this kind of datasets.
    \begin{itemize}
        \item \textbf{Real-world:} This is prepared from a real-world dataset by including missing values randomly. There is only one such dataset, namely Sporulation, which contains expression values of yeast genes measured across seven time points (0, 0.5, 2, 5, 7, 9, and 11.5h) during the sporulation process of budding yeast [12].
    \end{itemize}
\end{itemize}

\begin{table}[htp]
    \caption{Statistical and background details of the various dataseets used. Some line segments from some real-world datasets were cropped out to allow forming a realistic cluster.}
    \label{Table:Datasets}
    \centering
    \begin{tabular}{cccc}
        \hline
        \textbf{Type} & \textbf{Name} & \textbf{\# Data Points} & \textbf{Source} \\\hline
        Lines & Convex & 150 & Program-generated \\
        (Synthetic) & Doughnut & 400 & Program-generated \\\hline
         & Sparse Tripod & 50 & amtrak-routes [11] \\
        Lines & Dense Tripod & 100 & amtrak-routes [11] \\
        (Real-world) & Broken Beads 1 & 59 & Amtrak [11] \\
         & Broken Beads 2 & 50 & Amtrak [11] \\
         & Spectral Band 1 & 190 & Subway Lines [11] \\
         & Spectral Band 2 & 2000 & Subway Lines [11] \\\hline
        Points with Missing Entries & \multirow{2}{*}{Sporulation} & \multirow{2}{*}{475} & \multirow{2}{*}{Chu et al. [12]} \\
        (Real-world)  \\\hline
    \end{tabular}
\end{table}

\subsection{Empirical Analysis}
A set of programs are written in Python for the empirical analysis. All the experiments have been performed on an HP Laptop with 2.2 GHz Intel 13th Generation Core i7 processor, 18 MB Cache memory, and having 16 GB primary memory.

\subsubsection{Results on Line Datasets}
\begin{figure}[htp]
    \centering
    \begin{tabular}{ccc}
        \includegraphics[height=1.2in, width=1.4in]{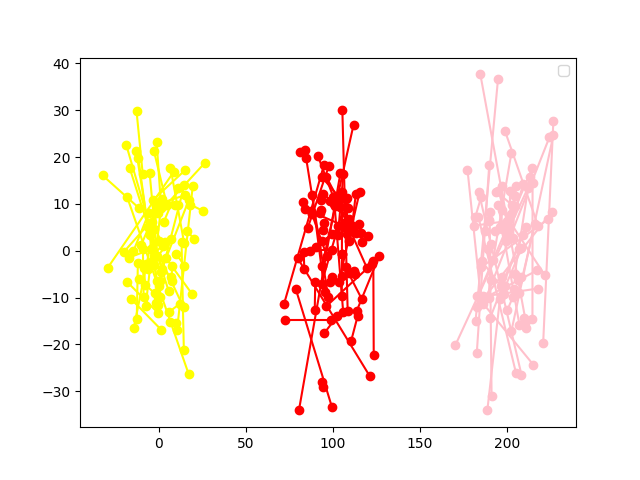} & \includegraphics[height=1.3in, width=1.4in]{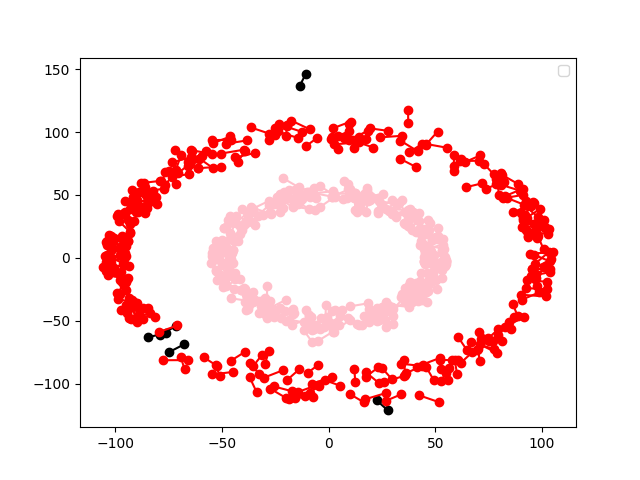} & \includegraphics[height=1.3in, width=1.4in]{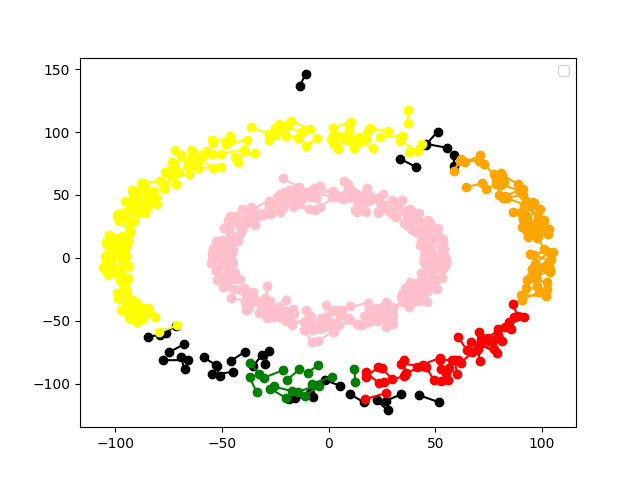} \\
        (a) & (b) & (c) \\
        \includegraphics[height=1.2in, width=1.4in]{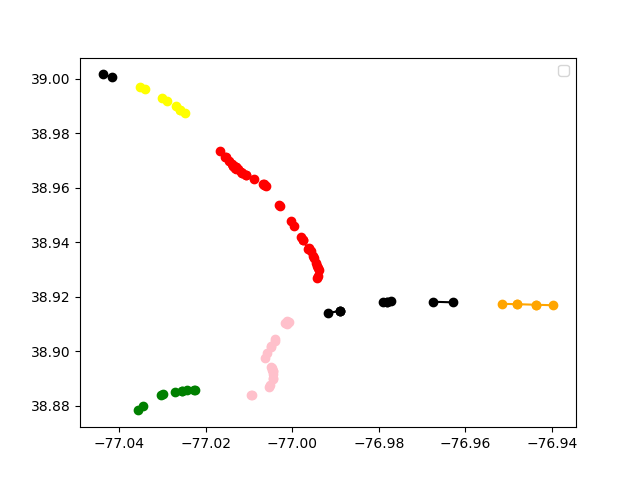} & \includegraphics[height=1.2in, width=1.4in]{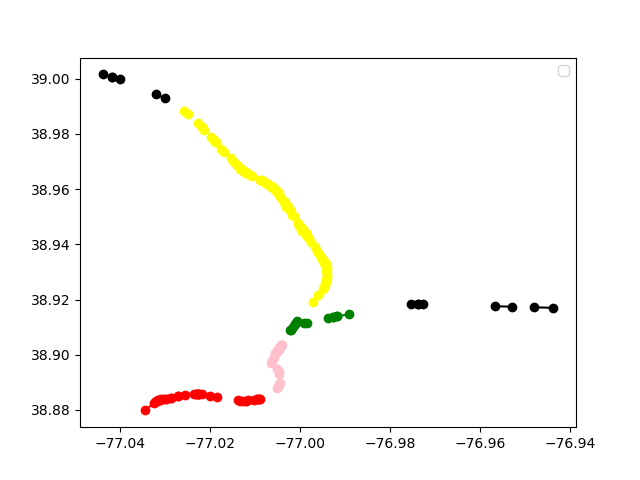} & \includegraphics[height=1.2in, width=1.4in]{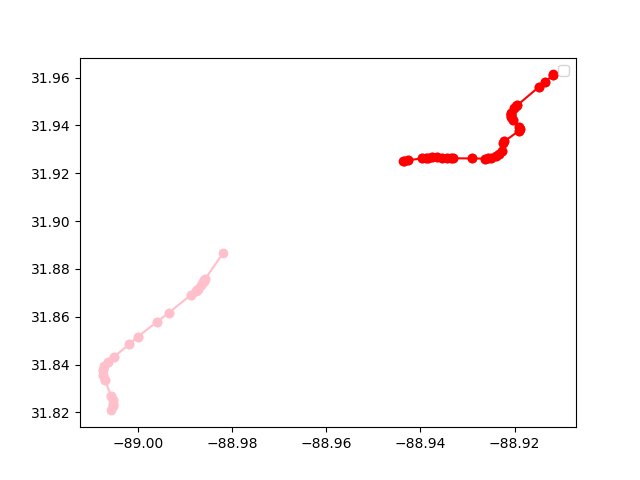} \\
        (d) & (e) & (f) \\
        \includegraphics[height=1.2in, width=1.4in]{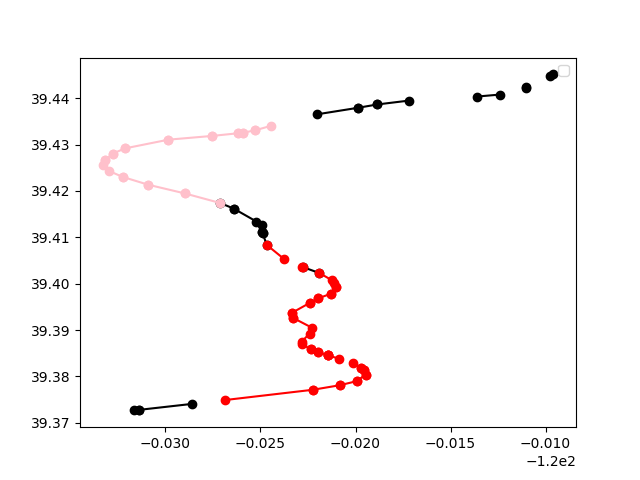} & \includegraphics[height=1.2in, width=1.4in]{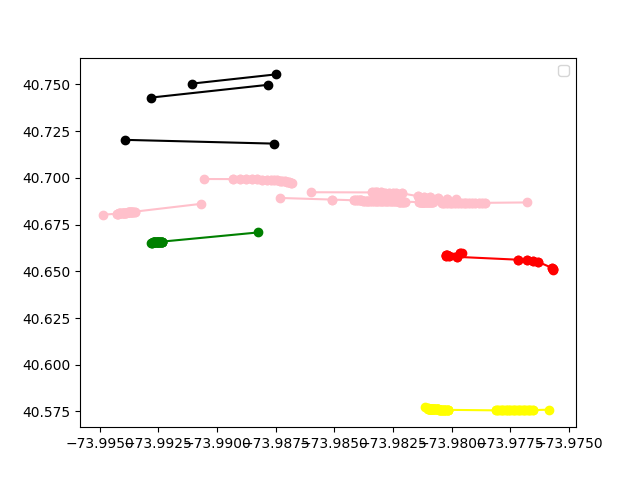} & \includegraphics[height=1.7in, width=2.5in]{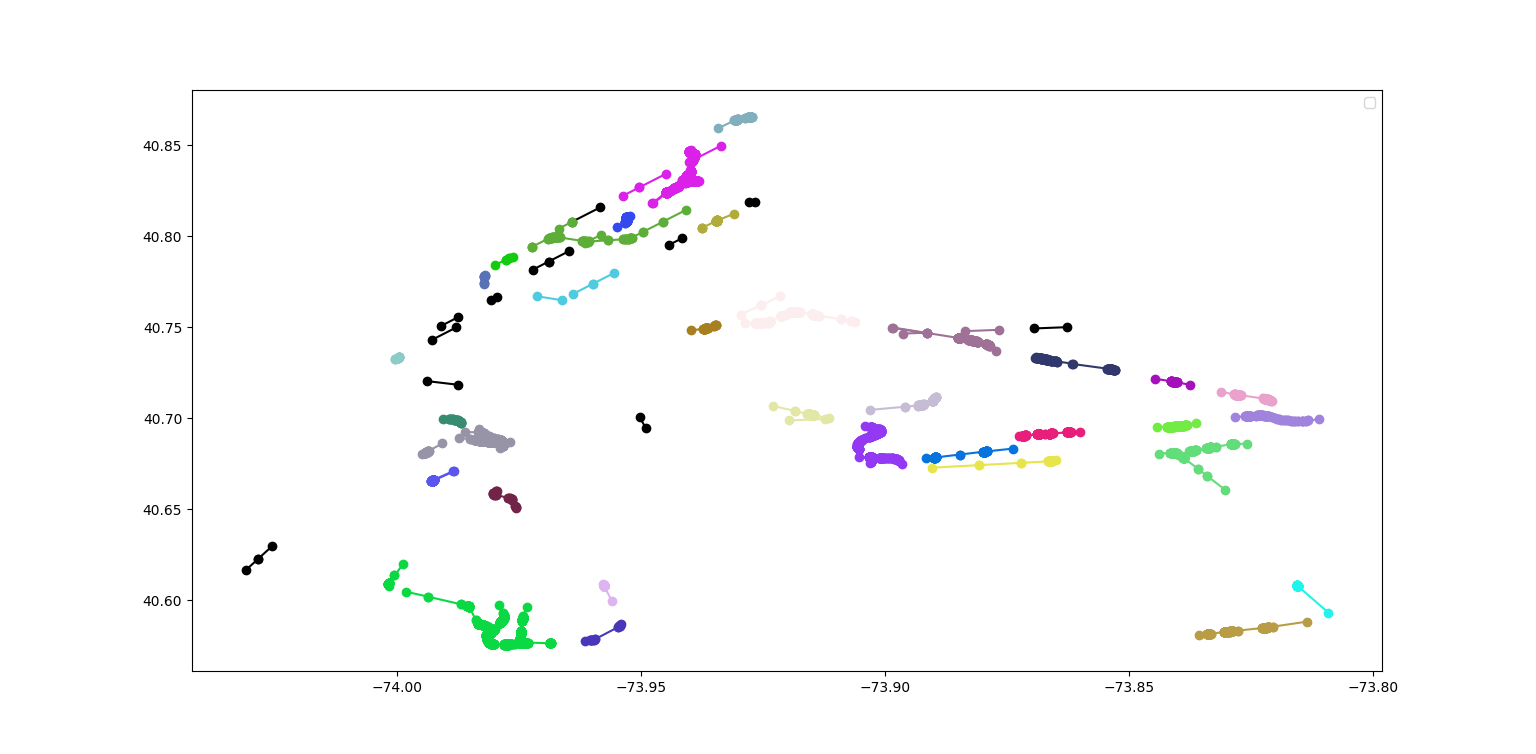} \\
        (g) & (h) & (i) \\
    \end{tabular}
    \caption{Clustering results for (a) Convex dataset considering $\alpha_l = 12$ and $c = 10$, (b) Doughnut dataset considering $\alpha_l = 12$ and $c = 5$, (c) Doughnut dataset considering $\alpha_l = 12$ and $c = 8$, (d) Sparse Tripod dataset considering $\alpha_l = 0.008 $ and $c = 3$, (e) Dense Tripod dataset considering $\alpha_l = 0.005$ and $c = 3$, (f) Broken Beads 1 dataset considering $\alpha_l = 0.005 $ and $c = 3$, (g) Broken Beads 2 dataset considering $\alpha_l = 0.005$ and $c = 8$, (h) Spectral Band 1 dataset considering $\alpha_l = 0.008$ and $c = 5$, (i) Spectral Band 2 dataset considering $\alpha_l = 0.005$ and $c = 3$.}
    \label{Figure:Line}
\end{figure}

The results obtained for the different line datasets are reported in Fig.~\ref{Figure:Line}. Notably, we apply only the first version of neighborhood computation of Algorithm~\ref{Algorithm:1} in all these cases due to the absence of domain knowledge. Moreover, as all the line segments have equal weightage, we take the scaling factor ($\alpha_l$) to be independent of the line segments. As can be seen from Figs.~\ref{Figure:Line}(a) and (b), the proposed approach is capable of exploring groups from both the convex and non-convex (with a little abuse of notation) orientation of lines. However, the cardinality parameter plays a crucial role in capturing the essence of true clusters in non-convex cases. For example, lowering the cardinality threshold of neighbourhood ($c$), helps to get larger clusters by ignoring small variations in density while increasing it, makes the algorithm sensitive to variation in density (compare Fig.~\ref{Figure:Line}(b) with Fig.~\ref{Figure:Line}(c)). The identification of outliers is also dependent on the cardinality parameter (more outliers in Fig.~\ref{Figure:Line}(c) than Fig.~\ref{Figure:Line}(b)). Note that the scaling factors ($\alpha_l$) were the same for the synthetic datasets because they were generated on a similar scale. However, for the other line datasets prepared from real-word data, we need to set the scaling factor ($\alpha_l$) separately. On carefully examining the real-word datasets (Figs.~\ref{Figure:Line}(d) and (e)), we can see that the algorithm can recognise dense and sparse regions along with outliers.  

\subsubsection{Results on Point Datasets with Missing Entries}
Clustering results are obtained for the Sporulation dataset (that we prepared by adding missing entries) considering $\alpha_l = 0.6$ and $c = 7$. The domain knowledge was incorporated to generate the neighbourhood by setting up an appropriate probability distribution. To be precise, we set the function $f_l$ as Unif[-4, 4] based the range of the values in respective columns in the dataset. The output yielded 5 clusters of yeast genes from the dataset along with a set of 120 outlier genes. This was similar to the results obtained with the original dataset in which no missing values were present. A detailed comparison of results is reported in the Table~\ref{Table:Comparison}. The same algorithm (with the same parameters) applied on the original dataset yielded 4 clusters of yeast genes from the dataset along with a set of 128 outlier genes. There was a significant overlap between the outliers identified from the original dataset and the dataset prepared by introducing missing entries. This demonstrates the effectiveness of the proposed approach when domain knowledge is brought in the model.

\begin{table}[htp]
  \caption{Comparison of clustering results obtained by applying the proposed algorithm on the original dataset and the dataset prepared by introducing missing entries.}
  \label{Table:Comparison}
  \centering
  \begin{tabular}{ccccccccc}
    \toprule
    \cmidrule(l){1-3}
    \textbf{Dataset} & \textbf{Algorithm} & \multicolumn{3}{c}{\textbf{Parameters}} & \multicolumn{4}{c}{\textbf{Clustering Results}} \\\cline{3-4}\cline{5-9}
     & \textbf{Version} & \textbf{$\alpha_l$} & \textbf{$c$} & \textbf{$f_l$} & \textbf{Count} & \textbf{Min} & \textbf{Max} & \textbf{\#Outliers} \\
    \midrule
    Sporulation & - & 0.6 & 7 & - & 4 & 17 & 219 & 128 \\
    (Original) &  &  &  &  &  &  &  &  \\\hline
    Sporulation & 3 & 0.6 & 7 & Unif[-4, 4] & 5 & 5 & 222 & 120 \\
    (With Missing Entries) &  &  &  &  &  &  &  \\
    \bottomrule
  \end{tabular}
\end{table}

\section{Conclusion and Future Work}
This paper presents a density based clustering algorithm that is particularly useful for grouping $n$-dimensional data points with missing entries. This approach does not require a pre-defined number of clusters. Moreover, it identifies outliers as noises. Interestingly, it can find arbitrarily-sized and arbitrarily-shaped clusters quite well. But this algorithm is not deterministic is nature. Through empirical analysis, we observe that it is effective on synthetic and real-world datasets. The proposed approach considers a number of parameters like volume, cardinality, and the density function to probabilistically generate the neighbourhood of a line. To fix the said parameters (or tune them easily), we require appropriate domain knowledge for the efficacy of the algorithm. Note that the current paper presents a simple implementation of the algorithm. If this is implemented using a range query approach with an appropriately maintained database index, we can obtain a better performance. The proposed clustering algorithm is capable of dealing with data that has at most one missing value per line. This approach can be further extended to the scenarios having higher number of missing entries per line.

\section*{References}
{
\small

\begin{enumerate}

    \item Duda, R. O., \& Hart, P. E. (2006) {\it Pattern classification}. John Wiley \& Sons.

    \item Marom, Y. (2019) k-Clustering of Lines and Its Applications. {\it Masters' Dissertation}, Department of Computer Science, University of Haifa.

    \item Marom, Y., \& Feldman, D. (2019) k-Means clustering of lines for big data. In {\it Advances in Neural Information Processing Systems}, 32.

    \item Ikotun, A. M., Ezugwu, A. E., Abualigah, L., Abuhaija, B., \& Heming, J. (2023) K-means clustering algorithms: A comprehensive review, variants analysis, and advances in the era of big data. {\it Information Sciences}, 622, 178-210.

    \item Peretz, T. (2011) Clustering of lines. {\it Masters' Dissertation}, Department of Mathematics and Computer Science, Open University of Israel.

    \item Ester, M., Kriegel, H. P., Sander, J., \& Xu, X. (1996) A density-based algorithm for discovering clusters in large spatial databases with noise. In {\it Proceedings of the Second International Conference on Knowledge Discovery and Data Mining}, Vol. 96, No. 34, pp. 226-231.

    \item Allison, P. D. (2009) Missing data. {\it The SAGE Handbook of Quantitative Methods in Psychology}, 72-89.

    \item Wilson, S. E. (2015) Methods for clustering data with missing values.

    \item Zass, R., \& Shashua, A. (2005) A unifying approach to hard and probabilistic clustering. In {\it Tenth IEEE International Conference on Computer Vision}, vol. 1, pp. 294-301.

    \item Stylianopoulos, K., \& Koutroumbas, K. (2021) A probabilistic clustering approach for detecting linear structures in two-dimensional spaces. {\it Pattern Recognition and Image Analysis}, 31:671-687.

    \item Line Datasets -- data.world. Accessed in May, 2024. Link: \url{https://data.world/datasets/line}

    \item Chu, S., DeRisi, J., Eisen, M., Mulholland, J., Botstein, D., Brown, P. O., \& Herskowitz, I. (1998) The transcriptional program of sporulation in budding yeast. {\it Science}, 282(5389), 699-705.

\end{enumerate}
}

\end{document}